\documentclass[12pt,letterpaper]{article}
\usepackage{situated}
\usepackage{latexsym,amssymb,amsmath}
\usepackage[breaklinks, colorlinks, linkcolor=black, urlcolor=black, citecolor=black]{hyperref}
\usepackage{path}
\usepackage{url}
\usepackage{verbatim}
\usepackage[pdftex]{graphicx}
\usepackage[round]{natbib}
\usepackage[usenames,dvipsnames]{xcolor}
\usepackage{microtype}

\usepackage[margin=1.09in]{geometry}

\numberwithin{equation}{section}
\numberwithin{table}{section}
\numberwithin{figure}{section}

\title{A Paradigm for Situated and Goal-Driven Language Learning}
\author{
Jon Gauthier$^{1,2}$ \\ \texttt{jon@gauthiers.net}
\And
Igor Mordatch$^{1,3}$ \\ \texttt{mordatch@openai.com}
\AND\\[-3ex]
{$^1$OpenAI\quad
$^2$Stanford NLP Group\quad
$^3$UC Berkeley}}

\begin{document}

%-------------------------Starts-------------------------------

\maketitle

A distinguishing property of human intelligence is the ability to flexibly use language in order to communicate complex ideas with other humans in a variety of contexts. Research in natural language dialogue should focus on designing communicative agents which can integrate themselves into these contexts and productively collaborate with humans. % Agents able to communicate and cooperate with us in our own language in real-world environments will have an impact far exceeding that of the non-verbal computer interfaces which have already greatly increased our productivity.

In this abstract, we propose a general situated language learning paradigm which is designed to bring about robust language agents able to cooperate productively with humans. This dialogue paradigm is built on a \textbf{utilitarian} definition of language understanding. Language is one of multiple tools which an agent may use to accomplish goals in its environment. We say an agent ``understands'' language only when it is able to use language productively to accomplish these goals. Under this definition, an agent's communication success reduces to its success on tasks within its environment.

This setup contrasts with many conventional natural language tasks, which maximize linguistic objectives derived from static datasets. Such applications often make the mistake of \emph{reifying language} as an end in itself. The tasks prioritize an isolated measure of linguistic intelligence (often one of linguistic competence, in the sense of \citet{chomsky1965aspects}), rather than measuring a model's effectiveness in real-world scenarios.\footnote{Our motivation here is similar to that of \citet{dagan2006pascal}, who recognized the negative effects of a community fragmented across different isolated application-specific tasks, and suggested the unified task of recognizing textual entailment (RTE) as a solution.} Our utilitarian definition is motivated by recent successes in reinforcement learning methods. In a reinforcement learning setting, agents maximize success metrics on real-world tasks, without requiring direct supervision of linguistic behavior.

\subsection*{The environment}

We propose an end-to-end learning environment with multiple language-enabled agents, each with the capacity to define their own internal goals and plans to reach those goals. Each agent may also have different capacities to observe or act in this environment. Their goals are grounded non-linguistic objectives: for example, to reach a desired location, manipulate objects in the environment, or transmit a piece of information.\footnote{A related line of work in evolutionary linguistics constructs a similar language learning scenario entirely without fixed-language agents \citep{smith2003iterated,steels_grounding_2012,kirby2014iterated}. All of the agents in these environments construct a novel language simultaneously to accomplish some shared task. This is an interesting separate line of research, but ultimately a separate task from the understanding and acquisition problems discussed in this abstract.} (We address the issue of non-linguistic grounding at length in the next section.) Some \emph{fixed-language agents} in the environment speak an existing conventional language (e.g. English), and other \emph{learning agents} are tasked with jointly learning this language while also solving other goals in the environment. The agents are assigned difficult (possibly distinct) tasks which require them to cooperate by communicating via their language channel.\footnote{
Much of the recent novel work in dialogue-based learning \citep{fleischman2005intentional,vogel2014learning,wang2016learning,weston2016dialog} and multi-agent communication \citep{lazaridou_towards_2016,andreas_reasoning_2016,foerster_learning_2016,sukhbaatar_learning_2016} can be fit into the paradigm described so far. This paper is concerned with designing a general paradigm for language-learning which contains these experiments, and picking out properties of the learning environment which are important for future research.}

In the simplest instance of the paradigm above, a single fixed-language \emph{parent} agent interacts with a learning \emph{child} agent. The parent cooperates with the child, but only when prompted through language. As a result, the child acquires the language of the parent in order to accomplish its task. This specific scenario is designed to train single communicative agents which can accomplish tasks in their environments with the help of human or simulated counterparts.

It is possible to include fixed-language agents in these environments without requiring human involvement. We have developed several instantiations of this paradigm in which the fixed-language agents speak some simplified hard-coded English or an artificial language (e.g. a programming language or query language). Near the limit of artificial complexity, agents might take advantage of the Internet to access world knowledge and synthesize responses. As our algorithms for learning in these environments improve, however, we expect that involving human agents in the environment would be an extremely effective way to train language-enabled artificial agents.

% Actually, this whole paragraph might be almost-redundant. Remove to save space? ~ JG
%
% The learning agents may also be allowed to interact with one another via language. This scenario may be useful to teach multiple agents with different skill sets (e.g., robots of different shapes and sizes) to cooperate by communicating.
% \JG{I removed the ``regularizing'' note; I think it doesn't affect the main message, and is only distracting..}
% Such interaction can have the effect of “regularizing” the language use of the learning agents, as noted by \citet{kirby2008}.

\subsection*{Environment grounding}

Our language-learning agents crucially need to be grounded in a world which is not only linguistic. This grounding may be physical --- for example, agents which are embodied in the real world --- or virtual. This grounding is what allows us to evaluate the agents in some way that does not prioritize language as the only objective. Grounding imposes additional responsibilities on the agent, such as competently sensing and acting in the world, but this should be seen as an opportunity rather than a problem. As we discuss in the following paragraphs, a grounded agent may use its experience in its environment to build better models of language, and likewise use its language to better reason about its environment.

This grounded environment is designed to bring about agents with comprehensive predictive models of the world, which combine linguistic knowledge with more general intelligent behavior. By design, we do not separate the activity of language model construction from many other intelligent predictive activities --- whether physical (predicting physical behavior of objects), psychological (modeling the beliefs and intentions of other agents), or social (understanding group membership and group-level action) \citep{lake16,clark13}. While early instantiations of this environment will limit the complexity necessary for the agents to model, we expect all of these different factors to eventually be relevant in a comprehensive learning environment.

\citet{mikolov_roadmap_2015} make a similar argument for grounding, but arrive at an environment in which perception and action is mediated only through a linguistic channel. While this textual interface certainly has the potential to simplify the problem, we believe there are several issues with this setup.

% First, by requiring that the agent interact with its environment only through text, we impose a particularly lossy representation of the world which may limit the agent's success. By definition, a textual representation must discretize and categorize observations into sequences of of atomic symbols. The pitfalls of such pre-specified representations have long been argued \citep{brooks91}. The advantage of avoiding these lossy formats has been further demonstrated by recent successes in end-to-end visuomotor approaches from game playing to robotics \citep{mnih15, levine15} which bypass intermediate representations, directly sensing camera pixels and outputting motor commands.

Textual interfaces impose a lossy representation of an agent's actual environment \citep{brooks91}, and necessarily bias the focus of an agent's observations. Consider the following text-only interaction in the style of \citet{mikolov_roadmap_2015} between a learning agent \emph{A} and omniscient agent \emph{B}, situated in some simulated physical world:

\begingroup
\fontsize{10pt}{12pt}\selectfont
\begin{verbatim}
    B: You are carrying a box of eggs and need to set them down.
    A: Is there a table nearby?
    B: There is a table to your left.
    A: I put the box on the table.
    B: The box slides off of the table and the eggs break open.
    [The table is missing a leg and tilted when the box was laid down.]
\end{verbatim}
\endgroup

The agent could have avoided this disaster if it had queried to find out that the table was missing a leg before placing weight on it. But we cannot expect an agent to exhaustively query its environment via text in general. (Should it also ask whether the table is made of solid material, or whether the table is on fire?) By contrast, a lossless visual observation does not remove possibly relevant information about the properties of tables, and it is up to the agent to learn what to attend to in such an environment and how to interpret these visual percepts.

For environments involving pre-programmed fixed-language agents, generating textual environment descriptions also places a significant implementational burden. It is relatively easy to generate physical motion and visual renderings of three-legged tables or toppling events, but much more difficult to describe these concepts in natural language.

Of course, even if we claim that text data is not enough, we must settle for some level of abstraction. A similar argument to the one above could be made for including sound-wave speech data in the agent's observations rather than only providing text and visual data. While this argument is likely also correct, we have to settle for a level of representation which is computationally tractable. Given that we want to make near-term progress on language comprehension and production, it seems reasonable to work with messages in text representation for the time being. It is likewise tractable to work in a virtual environment with simulated physics and visual inputs.

\subsection*{Conclusion}

In this abstract, we outlined a paradigm for grounded and goal-driven language learning in artificial agents. The paradigm is centered around a utilitarian definition of language understanding, which equates language understanding with the ability to cooperate with other language users in real-world environments. This position demotes language from its position as a separate task to be solved to one of several communicative tools agents might use to accomplish their real-world goals.

While this paradigm does already capture a small amount of recent work in dialogue, on the whole it has not received the focus it deserves in the research communities of natural language processing and machine learning. We hope this paper brings focus to the task of situated language learning as a way forward for research in natural language dialogue.

\subsection*{Acknowledgments}

The authors wish to thank their colleagues at OpenAI and Stanford for their useful comments and critiques.
% The authors wish to thank Pieter Abbeel, Danqi Chen, Jack Clark, Kevin Clark, Mihail Eric, Jim Fan, Jonathan Ho, Andrej Karpathy, Urvashi Khandelwal, Christopher Manning, Will Monroe, Sebastian Schuster, Zain Shah, Ilya Sutskever, Sida Wang, Keenon Werling, and their colleagues at OpenAI and Stanford for their comments and critiques.

%-------------------------Ends-------------------------------
\bibliographystyle{plainnat}
\bibliography{situated}

\begin{thebibliography}{17}
\providecommand{\natexlab}[1]{#1}
\providecommand{\url}[1]{\texttt{#1}}
\expandafter\ifx\csname urlstyle\endcsname\relax
  \providecommand{\doi}[1]{doi: #1}\else
  \providecommand{\doi}{doi: \begingroup \urlstyle{rm}\Url}\fi

\bibitem[Andreas and Klein(2016)]{andreas_reasoning_2016}
Jacob Andreas and Dan Klein.
\newblock Reasoning {About} {Pragmatics} with {Neural} {Listeners} and
  {Speakers}.
\newblock \emph{arXiv:1604.00562 [cs]}, April 2016.
\newblock URL \url{http://arxiv.org/abs/1604.00562}.
\newblock arXiv: 1604.00562.

\bibitem[Brooks(1991)]{brooks91}
Rodney Brooks.
\newblock Intelligence without representation.
\newblock \emph{Artificial Intelligence}, 47:\penalty0 139--159, 1991.

\bibitem[Chomsky(1965)]{chomsky1965aspects}
Noam Chomsky.
\newblock \emph{Aspects of the Theory of Syntax}.
\newblock M.I.T. Press, 1965.

\bibitem[Clark(2013)]{clark13}
Andy Clark.
\newblock Whatever next? predictive brains, situated agents, and the future of
  cognitive science.
\newblock \emph{Behavioral and Brain Sciences}, 36\penalty0 (03):\penalty0
  181--204, 5 2013.
\newblock ISSN 1469-1825.
\newblock \doi{10.1017/S0140525X12000477}.
\newblock URL \url{http://journals.cambridge.org/article_S0140525X12000477}.

\bibitem[Dagan et~al.(2006)Dagan, Glickman, and Magnini]{dagan2006pascal}
Ido Dagan, Oren Glickman, and Bernardo Magnini.
\newblock The pascal recognising textual entailment challenge.
\newblock In \emph{Proceedings of the First International Conference on Machine
  Learning Challenges: Evaluating Predictive Uncertainty Visual Object
  Classification, and Recognizing Textual Entailment}, MLCW'05, pages 177--190,
  Berlin, Heidelberg, 2006. Springer-Verlag.
\newblock ISBN 3-540-33427-0, 978-3-540-33427-9.
\newblock \doi{10.1007/11736790_9}.
\newblock URL \url{http://dx.doi.org/10.1007/11736790_9}.

\bibitem[Fleischman and Roy(2005)]{fleischman2005intentional}
Michael Fleischman and Deb Roy.
\newblock \emph{Proceedings of the Ninth Conference on Computational Natural
  Language Learning (CoNLL-2005)}, chapter Intentional Context in Situated
  Natural Language Learning, pages 104--111.
\newblock Association for Computational Linguistics, 2005.
\newblock URL \url{http://aclweb.org/anthology/W05-0614}.

\bibitem[Foerster et~al.(2016)Foerster, Assael, de~Freitas, and
  Whiteson]{foerster_learning_2016}
Jakob~N. Foerster, Yannis~M. Assael, Nando de~Freitas, and Shimon Whiteson.
\newblock Learning to {Communicate} to {Solve} {Riddles} with {Deep}
  {Distributed} {Recurrent} {Q}-{Networks}.
\newblock \emph{arXiv:1602.02672 [cs]}, February 2016.
\newblock URL \url{http://arxiv.org/abs/1602.02672}.
\newblock arXiv: 1602.02672.

\bibitem[Kirby et~al.(2014)Kirby, Griffiths, and Smith]{kirby2014iterated}
Simon Kirby, Tom Griffiths, and Kenny Smith.
\newblock Iterated learning and the evolution of language.
\newblock \emph{Current Opinion in Neurobiology}, 28:\penalty0 108 -- 114,
  2014.
\newblock ISSN 0959-4388.
\newblock \doi{http://dx.doi.org/10.1016/j.conb.2014.07.014}.
\newblock URL
  \url{http://www.sciencedirect.com/science/article/pii/S0959438814001421}.
\newblock SI: Communication and language.

\bibitem[Lake et~al.(2016)Lake, Ullman, Tenenbaum, and Gershman]{lake16}
Brenden~M. Lake, Tomer~D. Ullman, Joshua~B. Tenenbaum, and Samuel~J. Gershman.
\newblock Building machines that learn and think like people.
\newblock \emph{CoRR}, abs/1604.00289, 2016.
\newblock URL \url{http://arxiv.org/abs/1604.00289}.

\bibitem[Lazaridou et~al.(2016)Lazaridou, Pham, and
  Baroni]{lazaridou_towards_2016}
Angeliki Lazaridou, Nghia~The Pham, and Marco Baroni.
\newblock Towards {Multi}-{Agent} {Communication}-{Based} {Language}
  {Learning}.
\newblock \emph{arXiv:1605.07133 [cs]}, May 2016.
\newblock URL \url{http://arxiv.org/abs/1605.07133}.
\newblock arXiv: 1605.07133.

\bibitem[Mikolov et~al.(2015)Mikolov, Joulin, and Baroni]{mikolov_roadmap_2015}
Tomas Mikolov, Armand Joulin, and Marco Baroni.
\newblock A {Roadmap} towards {Machine} {Intelligence}.
\newblock \emph{arXiv:1511.08130 [cs]}, November 2015.
\newblock URL \url{http://arxiv.org/abs/1511.08130}.
\newblock arXiv: 1511.08130.

\bibitem[Smith et~al.(2003)Smith, Kirby, and Brighton]{smith2003iterated}
Kenneth Smith, Simon Kirby, and Henry Brighton.
\newblock Iterated learning: A framework for the emergence of language.
\newblock \emph{Artificial Life}, 9\penalty0 (4):\penalty0 371--386, 10 2003.
\newblock ISSN 1064-5462.
\newblock \doi{10.1162/106454603322694825}.
\newblock doi: 10.1162/106454603322694825.

\bibitem[Steels(2012)]{steels_grounding_2012}
Luc Steels.
\newblock Grounding language through evolutionary language games.
\newblock In \emph{Language {Grounding} in {Robots}}, pages 1--22. Springer,
  2012.
\newblock URL
  \url{http://link.springer.com/chapter/10.1007/978-1-4614-3064-3_1}.

\bibitem[Sukhbaatar et~al.(2016)Sukhbaatar, Szlam, and
  Fergus]{sukhbaatar_learning_2016}
Sainbayar Sukhbaatar, Arthur Szlam, and Rob Fergus.
\newblock Learning {Multiagent} {Communication} with {Backpropagation}.
\newblock \emph{arXiv:1605.07736 [cs]}, May 2016.
\newblock URL \url{http://arxiv.org/abs/1605.07736}.
\newblock arXiv: 1605.07736.

\bibitem[Vogel et~al.(2014)Vogel, G{\'o}mez~Emilsson, Frank, Jurafsky, and
  Potts]{vogel2014learning}
Adam Vogel, Andr{\'e}s G{\'o}mez~Emilsson, Michael~C. Frank, Dan Jurafsky, and
  Christopher Potts.
\newblock Learning to reason pragmatically with cognitive limitations.
\newblock In \emph{Proceedings of the 36th Annual Meeting of the {C}ognitive
  {S}cience {S}ociety}, pages 3055--3060, Wheat Ridge, CO, July 2014. Cognitive
  Science Society.

\bibitem[Wang et~al.(2016)Wang, Liang, and Manning]{wang2016learning}
Sida~I. Wang, Percy Liang, and Christopher~D. Manning.
\newblock Learning language games through interaction.
\newblock In \emph{Proceedings of the 54th Annual Meeting of the Association
  for Computational Linguistics (Volume 1: Long Papers)}, pages 2368--2378,
  Berlin, Germany, August 2016. Association for Computational Linguistics.
\newblock URL \url{http://www.aclweb.org/anthology/P16-1224}.

\bibitem[Weston(2016)]{weston2016dialog}
Jason Weston.
\newblock Dialog-based language learning.
\newblock \emph{arXiv preprint arXiv:1604.06045}, 2016.

\end{thebibliography}

\end{document}